\documentclass[sigconf]{acmart}
\usepackage{amsmath}
\usepackage{amsfonts}
\usepackage{makecell}
\usepackage{multirow}
\usepackage[caption=false, font=footnotesize]{subfig}
\usepackage{color}
\usepackage{algorithm}
\usepackage{algorithmic}
\usepackage{newfloat}
\usepackage{listings}

\newcommand{\shortModel}{Seq-HGNN}

\AtBeginDocument{%
  \providecommand\BibTeX{{%
    \normalfont B\kern-0.5em{\scshape i\kern-0.25em b}\kern-0.8em\TeX}}}

\setcopyright{acmlicensed}
\copyrightyear{2023}
\acmYear{2023}

\acmConference[SIGIR '23]{Proceedings of the 46th International ACM SIGIR Conference on Research and Development in Information Retrieval}{July      23--27, 2023}{Taipei, Taiwan}
\acmBooktitle{Proceedings of the 46th International ACM SIGIR Conference on Research and Development in Information Retrieval (SIGIR '23), July      23--27, 2023, Taipei, Taiwan}
\acmPrice{15.00}
\acmDOI{10.1145/3539618.3591765}
\acmISBN{978-1-4503-9408-6/23/07}

\acmSubmissionID{7969}

\begin{document}

\title{Seq-HGNN: Learning Sequential Node Representation \\ on Heterogeneous Graph}

\author{Chenguang Du}
\email{duchenguang@buaa.edu.com}
\affiliation{%
  \institution{SKLSDE, School of Computer Science, \\Beihang University}
  \city{Beijing}
  \country{China}
  \postcode{100191}
}

\author{Kaichun Yao}
\email{yaokaichun@outlook.com}
\affiliation{%
  \institution{Institute of Software \\Chinese Academy of Sciences}
  \city{Beijing}
  \country{China}
  \postcode{100190}
}

\author{Hengshu Zhu}
\email{zhuhengshu@gmail.com}
\authornote{Deqing Wang and Hengshu Zhu are corresponding authors.}
\affiliation{%
  \institution{Career Science Lab, \\BOSS Zhipin}
  \city{Beijing}
  \country{China}
  \postcode{100020}
}

\author{Deqing Wang}
\email{dqwang@buaa.edu.com}
\authornotemark[1]
\affiliation{%
  \institution{SKLSDE, School of Computer Science, \\Beihang University}
  \city{Beijing}
  \country{China}
  \postcode{100191}
}

\author{Fuzhen Zhuang}
\email{zhuangfuzhen@buaa.edu.com}
\affiliation{%
  \institution{Institute of Artificial Intelligence \&\\  SKLSDE, School of Computer Science \\Beihang University}
  \city{Beijing}
  \country{China}
  \postcode{100191}
}

\author{Hui Xiong}
\email{xionghui@ust.hk}
\affiliation{%
  \institution{Artificial Intelligence Thrust, \\The Hong Kong University of Science and Technology (Guangzhou)}
  \city{Guangzhou}
  \country{China}
  \postcode{511458}
}

\renewcommand{\shortauthors}{Chenguang Du, et al.}
\renewcommand{\shorttitle}{Seq-HGNN: Learning Sequential Node Representation on Heterogeneous Graph}

\begin{abstract}
  Recent years have witnessed the rapid development of heterogeneous graph neural networks (HGNNs) in information retrieval (IR) applications. 
  Many existing HGNNs design a variety of tailor-made graph convolutions to capture structural and semantic information in heterogeneous graphs.
  However, existing HGNNs usually represent each node as a \textit{single} vector in the multi-layer graph convolution calculation, which makes the high-level graph convolution layer fail to distinguish information from different relations and different orders, resulting in the information loss in the message passing. 
  To this end, we propose a novel heterogeneous graph neural network with \textit{sequential} node representation, namely \shortModel. 
  To avoid the information loss caused by the single vector node representation, we first design a sequential node representation learning mechanism to represent each node as a sequence of meta-path representations during the node message passing.
  Then we propose a heterogeneous representation fusion module, empowering \shortModel~ to identify important meta-paths and aggregate their representations into a compact one.
  We conduct extensive experiments on four widely used datasets from Heterogeneous Graph Benchmark (HGB) and Open Graph Benchmark (OGB).
  Experimental results show that our proposed method outperforms state-of-the-art baselines in both accuracy and efficiency.
  The source code is available at \url{https://github.com/nobrowning/SEQ_HGNN}.
\end{abstract}





\maketitle

\section{Introduction}
\label{sec:intro}
Heterogeneous graph-structured data widely exists in the real world, such as social networks, academic networks, user interaction networks, etc.
In order to take advantage of the rich structural and semantic information in heterogeneous graphs, heterogeneous graph neural networks (HGNNs) have been increasingly used in information retrieval (IR) applications, ranging from search engines~\cite{search_engines_GuanJSWYC22,search_engines_ChenWC00P22,search_engines_Yang20} and recommendation systems~\cite{recommendation_Cai22,recommendation_Song22, recommendation_PangWSZWXCLP22} to question answering systems~\cite{QA_FengHS022, QA_GaoZWDC0022}.

\begin{figure*}
    \centering
    \includegraphics[width=1.8\columnwidth]{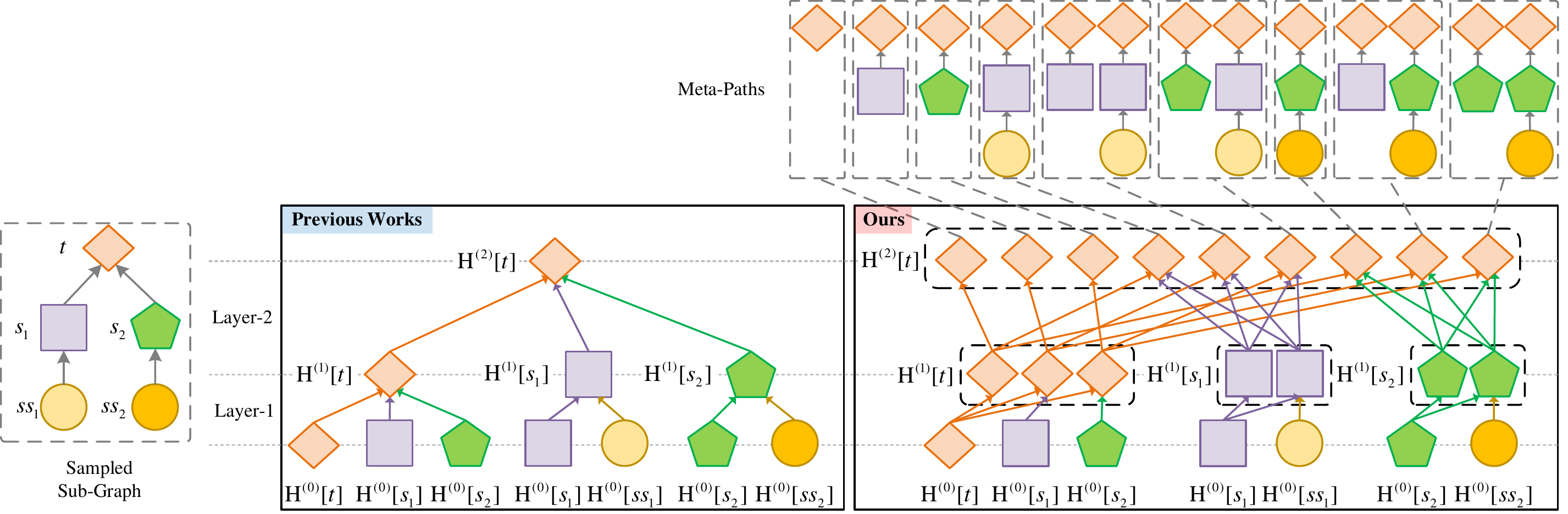}
    \caption{The comparison of node representation updates. The shapes of the nodes represent different node types.}
    \label{fig:node_update_compare}
    \Description{The comparison of node representation updates. The shapes of the nodes represent different node types.}
\end{figure*}

HGNNs can integrate structural and semantic information in heterogeneous graphs into node representations to meet downstream tasks. Existing HGNNs~\cite{rgcn, hgt, rhgnn} usually deploy multiple layers of graph convolution (i.e., message passing) to capture the neighborhood information of low-order and high-order neighbors in a graph. For a particular node, each layer of convolutions represents it as one single vector, which is the input of the next higher layer. Consequently, the single vector incorporates mixed neighbor information from different relationships and distinct orders. That is, higher-level convolutions are incapable of distinguishing messages from various sources by a single vector which leads to structural information loss and difficulty in refining message passing strategy.
Here we take a classic graph learning for instance, as shown in Figure~\ref{fig:node_update_compare}, a sampled sub-graph contains target node $t$, two source nodes ($s_1$ and $s_2$) and two 2-hop source nodes ($ss_1$ and $ss_2$). The $ss_1$ and $ss_2$ are the source node of ($s_1$ and $s_2$), respectively.
Existing methods~\cite{rgcn,hgt,rhgnn} usually conduct graph convolution operations twice to learn the node representation. Through the first layer of graph convolution, the target node $t$ and its neighbors (source nodes) are represented as $\mathbf{H}^{(1)}[t]$, $\mathbf{H}^{(1)}[s_1]$, and $\mathbf{H}^{(1)}[s_2]$, respectively, which are used as the input of the next layer of graph convolution computation. 
The information in $s_1$ and $ss_1$ is mixed in $\mathbf{H}^{(1)}[s_1]$ and information of $s_2$ and $ss_2$ is mixed in $\mathbf{H}^{(1)}[s_2]$. Based on the $\mathbf{H}^{(1)}[t]$, $\mathbf{H}^{(1)}[s_1]$, and $\mathbf{H}^{(1)}[s_2]$, the second layer of graph convolution cannot distinguish the information from $ss_1$ and $s_1$ and the information from $ss_2$ and $s_2$.

Intuitively, the semantics learned from each layer and each relation can reflect different-grained features, which strongly correlate to the different tasks, while the mixtures of all information may lead to sub-optimal results for the downstream tasks.

Along this line, we propose a novel heterogeneous graph neural network with \textit{sequential} node representation (\shortModel), which learns representations of meta-paths and fuses them into high-quality node representations.
Specifically, we first propose a sequential node representation learning mechanism that performs message passing over all meta-paths within fixed hops and represents each node as a sequence of meta-path representation.
As Figure~\ref{fig:node_update_compare} illustrates, after the calculation of two \shortModel~ layers, \shortModel~ can automatically capture the information of all meta-paths and their combinations within 2 hops, which are respectively stored in multiple independent vectors. These vectors then form a sequence as the representation of target $t$ (i.e. $\mathbf{H}^{(2)}[t]$).
The sequential representation enables higher \shortModel~ layers to naturally distinguish messages from different meta-paths.
Secondly, we design a heterogeneous representation fusion module to transform the sequence-based node representations into a compact representation, which can be used in various downstream tasks.
Also, \shortModel~ can benefit the discovery of effective entities and relations by estimating the importance of different meta-paths.
Finally, we conduct extensive experiments on real-world datasets. The experimental results show that \shortModel~ achieves the best performance compared with several state-of-the-art baselines.

Our contributions can be summarized as follows:
\begin{itemize}
    \item We propose a novel heterogeneous graph representation learning model with sequential node representation, namely \shortModel. To the best of our knowledge, the \shortModel~ is the first work to represent nodes as sequences, which can provide better representations by recording messages passing along multiple meta-paths intact.
    \item We conduct extensive experiments on four widely used datasets from Heterogeneous Graph Benchmark (HGB)~\cite{HGB} and Open Graph Benchmark (OGB)~\cite{ogb} to demonstrate the advantage of our model over state-of-the-art baselines.
    \item Our model performs good interpretability by analyzing the attention weight of meta-paths in heterogeneous graphs.
\end{itemize}

\section{Related Work}
In this section, we introduce the related work on heterogeneous graph neural networks and the applications of heterogeneous graph neural networks in the field of information retrieval.

\subsection{Heterogeneous graph neural networks}
Heterogeneous graph neural networks (HGNNs) are proposed to deal with heterogeneous graph data. Some HGNNs apply graph convolution directly on original heterogeneous graphs. RGCN~\cite{rgcn} is a widely-used HGNN, which sets different transfer matrices for different relations in heterogeneous graphs. R-HGNN~\cite{rhgnn} learned different node representations under each relation and fuses representations from different relations into a comprehensive representation. Other HGNNs used meta-paths to adopt homogeneous-graph-based methods on the heterogeneous graph. For instance, HAN~\cite{han} utilized GAT~\cite{gat} to calculate node-level and semantic-level attention on meta-path-based sub-graphs. MAGNN~\cite{magnn} introduced intra-meta-path aggregation and inter-meta-path aggregation to capture information on the heterogeneous graph. HeCo~\cite{heco} selected positive sample nodes based on meta-path on heterogeneous graph comparative learning. The meta-path-based methods require manual-designed meaningful meta-paths and can not be applied in large-scale heterogeneous graphs limited by the computational complexity~\cite{rhgnn}.
To overcome the disadvantages of meta-path, Het-SANN~\cite{Het-SANN} aggregated multi-relational information of projected nodes by attention-based averaging. GTN~\cite{gtn} and ie-HGCN~\cite{ie-hgcn} were designed to discover effective meta-paths for the target nodes. HGT~\cite{hgt} introduced the dot product attention mechanism~\cite{transformer} into heterogeneous graph learning, which can learn the implicit meta-paths. These methods represented each node as one single vector, which means confounding messages from different relations and orders, resulting in the loss of structural information.

In more recent years, in light of learning comprehensive node representations, some researchers adopted Simplified Graph Convolutional Network (SGC)~\cite{SGC}-based methods for heterogeneous graph processing~\cite{NARS, GAMLP, SeHGNN}. The core points of them focused on subgraph division and preprocessing. To be specific, these methods first divided a heterogeneous graph into several relation-driven subgraphs based and then conducted simple message passing and pre-computation in the preprocessing stage. However, there are two main drawbacks with this design making them unsuitable for application scenarios: Firstly, multiple downstream tasks are needed to meet the requirements of different messaging passing. For instance, in link prediction tasks, models need to mask some links in the graph, while using SGC-based methods means performing multiple separate preprocessing pipelines, resulting in high computational consumption for various downstream tasks. Secondly, SGC-based methods necessitate learning a distinct set of model parameters for each class of nodes in a heterogeneous graph, with no correlation between parameters of different node types. Such approaches lack the capacity for transfer learning across diverse node types. Specifically, the training and optimization of a particular node type in a heterogeneous graph using SGC-based methods do not contribute to performance enhancement in predicting other node types.

Unlike previous works, our model implements sequential node representation, which records messages from all meta-paths within a fixed step and achieves better performance and interpretability. Moreover, our model possesses end-to-end learning capabilities, enabling it to handle various downstream tasks with a more general and simplified workflow.

\subsection{HGNNs applications in IR}
In recent years, heterogeneous graph neural networks (HGNNs) have emerged as a powerful tool for extracting rich structural and semantic information from heterogeneous graphs, and have consequently found numerous applications in information retrieval (IR) domains.

In the realm of search engines and matching, \citet{search_engines_ChenWC00P22} proposed a cross-modal retrieval method using heterogeneous graph embeddings to preserve abundant cross-modal information, addressing the limitations of conventional methods that often lose modality-specific information in the process. \citet{search_engines_GuanJSWYC22} tackled the problem of fashion compatibility modeling by incorporating user preferences and attribute entities in their meta-path-guided heterogeneous graph learning approach.
\citet{poi} introduced the Spatio-Temporal Dual Graph Attention Network (STDGAT) for intelligent query-Point of Interest (POI) matching in location-based services, leveraging semantic representation, dual graph attention, and spatiotemporal factors to improve matching accuracy even with partial query keywords. \citet{pjfit} proposed a knowledge-enhanced person-job fit approach based on heterogeneous graph neural networks, which can use structural information to improve the matching accuracy of resumes and positions.

Recommendation systems have also benefited from HGNNs. \citet{recommendation_Cai22} presented an inductive heterogeneous graph neural network (IHGNN) model to address the sparsity of user attributes in cold-start recommendation systems. \citet{recommendation_PangWSZWXCLP22} proposed a personalized session-based recommendation method using heterogeneous global graph neural networks (HG-GNN) to capture user preferences from current and historical sessions. Additionally, \citet{recommendation_Song22} developed a self-supervised, calorie-aware heterogeneous graph network (SCHGN) for food recommendation, incorporating user preferences and ingredient relationships to enhance recommendations.

HGNNs have also garnered attention from scholars in the field of question-answering systems. For example, \citet{QA_FengHS022} proposed a document-entity heterogeneous graph network (DEHG) to integrate structured and unstructured information sources, enabling multi-hop reasoning for open-domain question answering. \citet{QA_GaoZWDC0022} introduced HeteroQA, which uses a question-aware heterogeneous graph transformer to incorporate multiple information sources from user communities.

\section{Preliminaries}

\textbf{Heterogeneous Graph:} Heterogeneous graph is defined as a directed graph $G = (V, E)$, with node type mapping $\tau: V\to A$ and edge type mapping $\phi: E \to R$, where $V$ is the node set, $E$ is the edge set, $A$ and $R$ represent the set of node types and edge types respectively, and $|A| + |R|> 2$. 

\textbf{Relation:} For an edge $e = (s, t)$ linked from source node $s$ to target node $t$, the corresponding relation is $r = <\tau(s), \phi(e), \tau(t)>$. A heterogeneous graph can be considered a collection of triples consisting of source nodes $s$ linked to the target nodes $t$ through edges $e$.

\textbf{Relational Bipartite Graph:} Given a heterogeneous graph $G$ and a relation $r$, the bipartite graph $G_{r}$ is defined as a graph composed of all the edges of the corresponding type of the relation $r$. In other words, $G_{r}$ contains all triples $<s, e, t>$, where the relation $\phi(e)=r$.

\textbf{Meta-path:} Meta-path $P$ is defined as a path with the following form: $A_1 \xrightarrow{r_1} A_2 \xrightarrow{r_2} \cdots \xrightarrow{r_{l-1}} A_l$ (abbreviated as $A_1 A_2 \cdots A_l$), where $A_i \in A, r_i \in R$. The meta-path describes a composite relation between node types $A_1$ and $A_l$, which expresses specific semantics.

\textbf{Graph Representation Learning:} Given a graph $G = (V, E)$, graph representation learning aims to learn a function $V \to \mathbb{R}^d, d \ll |V|$ to map the nodes in the graph to a low-dimensional vector space while preserving both the node features and the topological structure information of the graph. These node representation vectors can be used for a variety of downstream tasks, such as node classification and link prediction.

\begin{figure*}[t]
    \centering
    \includegraphics[width=2\columnwidth]{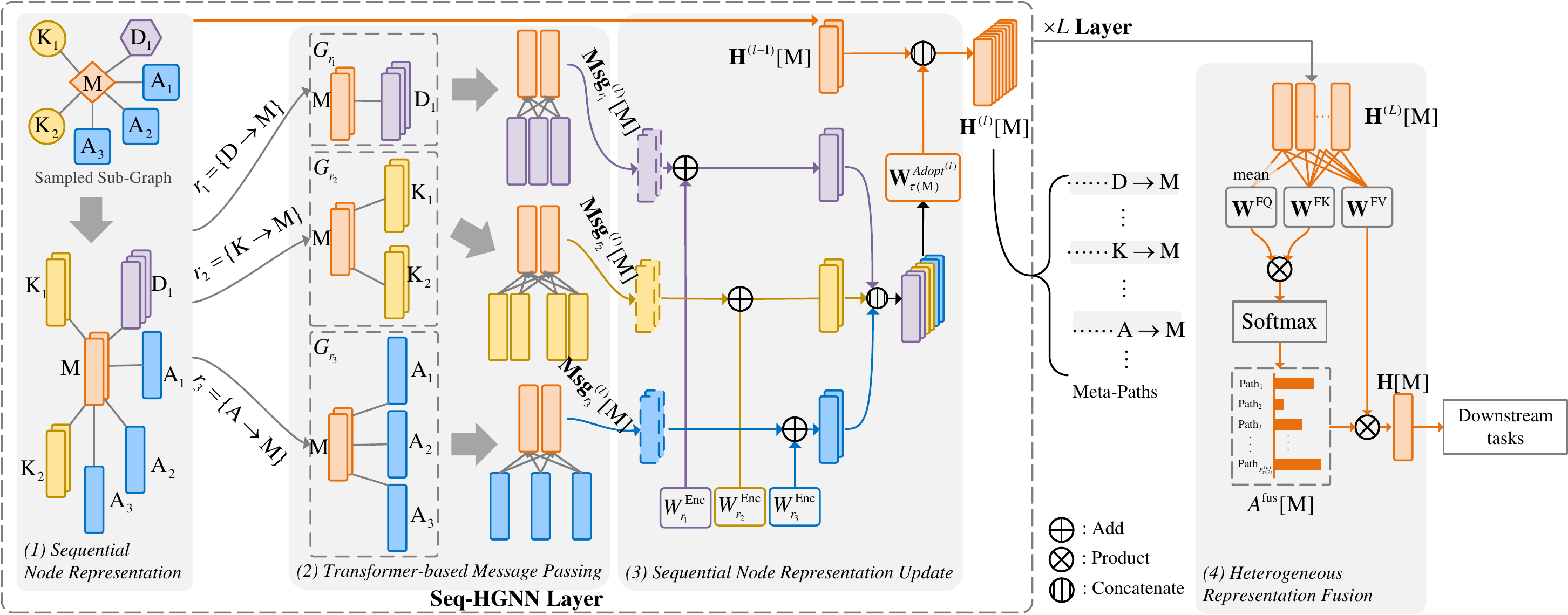}
    \caption{The overview of our proposed \shortModel. Given a heterogeneous sub-graph containing a target node $\mathbf{M}$ and six source nodes, \shortModel~ first learns a sequential node representation of $\mathbf{M}$ (i.e. $\mathbf{H}^{(L)}\left[\mathbf{M}\right]$), and then fuses the representation $\mathbf{H}^{(L)}\left[\mathbf{M}\right]$ for multiple downstream tasks. In the sub-graph, $\mathbf{M}$, $\mathbf{K}$, $\mathbf{A}$, and $\mathbf{D}$ represent node types \textit{Movie}, \textit{Keyword}, \textit{Actor}, \textit{Director}, respectively.}
    \label{fig:model}
    \Description{The overview of our proposed \shortModel. Given a heterogeneous sub-graph containing a target node $\mathbf{M}$ and six source nodes, \shortModel~ first learns a sequential node representation of $\mathbf{M}$ (i.e. $\mathbf{H}^{(L)}\left[\mathbf{M}\right]$), and then fuses the representation $\mathbf{H}^{(L)}\left[\mathbf{M}\right]$ for multiple downstream tasks. In the sub-graph, $\mathbf{M}$, $\mathbf{K}$, $\mathbf{A}$, and $\mathbf{D}$ represent node types \textit{Movie}, \textit{Keyword}, \textit{Actor}, \textit{Director}, respectively.}
\end{figure*}
\section{Methodology}
The overview of the proposed \shortModel~ is shown in Figure~\ref{fig:model}. The \shortModel~ is composed of multiple \textbf{\shortModel~ Layers} and a \textbf{Heterogeneous Representation Fusion} module.
The \shortModel~ Layers aggregate the information provided by the source node $s$, and update the representation of the target node $t$. We denote the output representation of the $l$-th layer as $\mathbf{H}^{(l)}$, which is also the input of the ($l+1$)-th layer ($1 \leq l \leq L$). By stacking $L$ \shortModel~ Layers, each target node $t$ can receive higher-order neighbor information.
The \shortModel~ Layer consists of three modules: \textit{Sequential Node Representation}, \textit{Transformer-based Message Passing} and \textit{Sequential Node Representation Update}. 
Among them, the \textit{Sequential Node Representation} transforms each node into a set of representation vectors.
The \textit{Transformer-based Message Passing} generates neighbor messages for the target node by aggregating the information of neighbors (source nodes).
The \textit{Sequential Node Representation Update} computes a new representation for $t$ based on the representation from the previous layer and the received neighbor messages.
Finally, the Heterogeneous Representation Fusion module estimates the importance of meta-paths and fuses the representations of meta-paths to a single vector as node representation, which can be utilized in downstream tasks.

\subsection{Sequential Node Representation}
\label{node_feature}
In heterogeneous graphs, the nodes often have multiple attributes and receive messages from multiple types of nodes. For example, in a heterogeneous graph from a movie review website, a \textit{Movie} node usually contains multiple description attributes such as Storyline, Taglines, Release date, etc. Existing methods only support representing each node as a single vector, which implies that the multiple properties of each node are confused into one vector. This causes information loss of node representation. 

Different from the above-mentioned graph representation learning methods~\cite{rgcn,hgt,rhgnn}, we represent each node as one sequence of vectors, which can record multiple properties of node and messages from multiple meta-paths intact.
Concretely, given a node $i$, we first design a type-specific transform matrix $W^{\tau(i)}$ to convert features $x^i$ of node $i$ to the same space:
\begin{equation}
    H^{(0)}_f\left[i\right] = W^{\tau(i)}_{f} \cdot x^i_f + b^{\tau(i)}_{f},
\end{equation}
where $\tau(i)$ is the node type of node $i$; $1 \leq f \leq F^{(0)}_{\tau(i)}$; $F^{(0)}_{\tau(i)}$ is the number of $i$'s features; $x^i_f$ is the $f$-th initialized feature in the feature sequence of $i$; $H^{(0)}_f\left[i\right] \in \mathbb{R}^d$ is the node features after the transform; $b^{\tau(i)}_{f}$ is the bias; $d$ is the dimension of features.

Next, we concatenate the $F^{(0)}_{\tau(i)}$ transformed representations of node $i$ to get an input sequence $\mathbf{H}^{(0)}\left[i\right]$ for the \shortModel~ model:
\begin{equation}
    \mathbf{H}^{(0)}\left[i\right] = \bigg\Arrowvert_{f}^{F^{(0)}_{\tau(i)}} H^{(0)}_f\left[i\right],
\end{equation}
where $\bigg\Arrowvert$ is the concatenation operation and $\mathbf{H}^{(0)}\left[i\right] \in \mathbb{R}^{F^{(0)}_{\tau(i)} \times d}$ is a sequence with the length of $F^{(0)}_{\tau(i)}$.

It is worth noting that our proposed sequential node representation is independent of time series.
During the message passing, our model always represents each node as one sequence of vectors. Each vector in the sequence can represent either the meta-path information or a specific feature attribute of the node.
For a detailed description, please refer to Section~\ref{message_pass} and~\ref{node_update}.

\subsection{Transformer-based Message Passing}
\label{message_pass}
The message-passing module aggregates the information of neighbors (source nodes) on each relational bipartite graph to generate neighbor messages for the target node.

\subsubsection{Neighbor Importance Estimation}
\label{msg_pass}
Before the neighbor message generation, we first estimate the importance of these neighbors.
We utilize the mutual attention~\cite{transformer,hgt} to calculate the importance of source nodes to the target node.
Specifically, we first project the representations of the target node $t$ and its neighbors (source nodes $s$) to multiple Query vectors $\mathbf{Q}$ and Key vectors $\mathbf{K}$, respectively.
\begin{equation}
    \label{equ:cal_q}
    \mathbf{Q}^{(l)}[t] = \bigg\Arrowvert_{f}^{F^{(l-1)}_{\tau(t)}} \mathbf{W}^{\text{Query}^{(l)}}_{\tau(t)}H^{(l-1)}_f\left[t\right] + b^{\text{Query}^{(l)}}_{\tau(t)},
\end{equation}
\begin{equation}
    \label{equ:cal_k}
    \mathbf{K}^{(l)}[s] = \bigg\Arrowvert_{f}^{F^{(l-1)}_{\tau(s)}} \mathbf{W}^{\text{Key}^{(l)}}_{\tau(s)}H^{(l-1)}_f\left[s\right] + b^{\text{Key}^{(l)}}_{\tau(s)},
\end{equation}
where $\mathbf{W}^{\text{Query}^{(l)}}_{\tau(t)} \in \mathbb{R}^{d \times d}$ and $\mathbf{W}^{\text{Key}^{(l)}}_{\tau(s)} \in \mathbb{R}^{d \times d}$ are type-specific trainable transformation matrices for source node $s$ and target node $t$; $b^{\text{Query}^{(l)}}_{\tau(t)}$ and $b^{\text{Key}^{(l)}}_{\tau(s)}$ are bias vectors. The shapes of $\mathbf{Q}^{(l)}[t]$ and $\mathbf{K}^{(l)}[s]$ are $F^{(l-1)}_{\tau(t)} \times d$ and $F^{(l-1)}_{\tau(s)} \times d$, respectively.
$F^{(l-1)}_{\tau(t)}$ and $F^{(l-1)}_{\tau(s)}$ represent the length of sequence representations of $t$ and $s$ in the ($l-1$) layer, respectively.

We regard the attention weights of the source node $s$ to the target node $t$ as the importance of $s$ to $t$.
Since the nodes would play different roles in different relations, we calculate the attention weights on each bipartite graph separately. More specifically, we denote the set of source nodes connected by the target node $t$ in the bipartite graph $G_{r}$ as $N_{r}(t)$, where $r \in \mathbf{R}$. Then, the attention weights can be formulated as:
\begin{equation}
    \textbf{Attn}^{(l)}_{r}[s,t] = \underset{\forall s \in N_{r}(t)}{\text{Softmax}} \left(  \mathbf{K}^{(l)}[s]W^{\text{ATT}^{(l)}}_{r} {\mathbf{Q}^{(l)}[t]}^\top \right) \cdot \frac{1}{\sqrt{d}},
\end{equation}
where $\textbf{Attn}^{(l)}_{r}[s,t]$ is the importance estimation of the source node $s$ to the target node $t$ on relation $r$, and $W^{\text{ATT}^{(l)}}_{r} \in \mathbb{R}^{d \times d}$ is the transform matrix for relation $r$.

Unlike the existing attention-based approaches~\cite{rhgnn,hgt,gat}, the attention weight $\textbf{Attn}^{(l)}_r[s,t]$ is a matrix with the shape $F^{(l-1 )}_{\tau(s)} \times F^{(l-1)}_{\tau(t)}$ rather than a scalar. 
Each element in $\textbf{Attn}^{(l)}_r[s,t]$ represents the attention weight of an item in the representation sequence of $s$ to an item in the representation sequence of $t$.

\subsubsection{Neighbor Message Generation}
According to the importance of neighbors, the \shortModel~ aggregates the neighbor information and treats it as the neighbor messages for $t$.

First, \shortModel~ extracts features of the source node $s$ in each bipartite graph $G_r$ separately as follows:
\begin{equation}
    \textbf{Ext}^{(l)}_{r}[s] = \bigg\Arrowvert_{f}^{F^{(l-1)}_{\tau(s)}} 
    W^{\text{EXT}^{(l)}}_{r} 
    \left(\mathbf{W}^{\text{Value}^{(l)}}_{\tau(s)}H^{(l-1)}_f\left[s\right] + b^{\text{Value}^{(l)}}_{\tau(s)}\right),
\end{equation}
where $\textbf{Ext}^{(l)}_{r}[s] \in \mathbb{R}^{F^{(l-1)}_{\tau(s)} \times d}$ is the extracted message from the source node $s$ under the relation $r$; $\mathbf{W}^{\text{Value}^{(l)}}_{\tau(s)} \in \mathbb{R}^{d \times d}$ is the transformation matrix for for the node type $\tau(s)$; $b^{\text{Value}^{(l)}}$ is the bias; $W^{\text{EXT}^{(l)}}_{r}$ is the transform matrix for the relation $r$.

Then, 
we can obtain the neighbor messages for $t$ under relation $r$ as follows:
\begin{equation}
    \textbf{Msg}^{(l)}_{r}[t] = \underset{\forall s \in N_{r}(t)}{\sum} \left( {\textbf{Attn}^{(l)}_{r}[s,t]}^\top \textbf{Ext}^{(l)}_{r}[s] \right) ,
\end{equation}
where $\textbf{Msg}^{(l)}_{r}[t] \in \mathbb{R}^{F^{(l-1)}_{\tau(t)} \times d}$ is a sequence with the same shape as the node representation $\mathbf{H}^{(l-1)}\left[t\right]$, and $N_{r}(t)$ is the set of neighbors (source nodes) of the target node $t$ in the bipartite graph $G_r$.

\subsection{Sequential Node Representation Update}
\label{node_update}
After the message passing process, the target node $t$ receives messages $\textbf{Msg}^{(l)}_{r}[t]$ from multiple relations $r \in R$. Based on the received messages and the representations from the previous layer $\mathbf{H}^{(l-1)}\left[t\right]$, we get the updated node representation of $t$.

First, we concatenate the message sequences from different relation types with relation-aware encoding as follows:

\begin{align}
    \widetilde{\mathbf{H}}^{(l)}\left[t\right] &= \underset{\forall r \in R(t)}{\Arrowvert}\widetilde{\textbf{Msg}}^{(l)}_{r}[t],\\
    \widetilde{\textbf{Msg}}^{(l)}_{r}[t] &= \textbf{Msg}^{(l)}_{r}[t] \oplus W_r^{\text{Enc}},
    \label{eq:rel_enc}
\end{align}
where $R(t)$ is the set of relation types whose target node type is $\tau(t)$; $W_r^{\text{Enc}} \in \mathbb{R}^{d}$ is the relation encoding for relation $r$, which is a learnable vector to distinguish messages from different relation types; $\oplus$ represents that the relation encoding is added to each vector in the sequence.

Then, we concatenate the representations of the target node from the last layer and encoded messages to obtain a new representation of the target node $t$: 
\begin{equation}
    \mathbf{H}^{(l)}\left[t\right] = \mathbf{H}^{(l-1)}\left[t\right] \quad \Arrowvert \quad \mathbf{W}^{\text{Adopt}^{(l)}}_{\tau(t)}\widetilde{\mathbf{H}}^{(l)}\left[t\right],
    \label{equ:node_update}
\end{equation}
where $\mathbf{H}^{(l)}\left[t\right] \in \mathbb{R}^{F^{(l)}_{\tau(t)} \times d}$ is the updated representations of target node $t$; $\mathbf{W}^{\text{Adopt}^{(l)}}_{\tau(t)} \in \mathbb{R}^{d \times d}$ is a transformation matrix corresponding to the $\tau(t)$.

We denote that the number of relation types connected to the target node $t$ is $\text{len}(R(t))$, then the length of the sequential representations for target node $t$ grows according to the following:
\begin{equation}
    F^{(l)}_{\tau(t)} = F^{(l-1)}_{\tau(t)} \times \left(\text{len}(R(t)) + 1 \right),
    \label{equ:seq_grow}
\end{equation}
where $F^{(l-1)}_{\tau(t)}$ and $F^{(l)}_{\tau(t)}$ represent the length of the sequential representation for node $t$ in the $(l-1)$-th and $l$-th layers, respectively.
Referring to Equation~\ref{equ:node_update} and Equation~\ref{equ:seq_grow}, we can summarize that in sequential node representation, information from a node itself and low-order neighbors is located at the beginning of the sequence, followed by high-order information. As deeper Seq-HGNN Layers are performed, information from higher-order neighbors is appended to the sequence.

\subsection{Heterogeneous Representation Fusion}
\label{sec:feat_fus}
After the $L$-layer \shortModel~ computation, each target node $t$ is represented by a sequence with length $F^{(L)}_{\tau(t)}$, which are the representations of the $t$ from multiple meta-paths.
We utilize the self attention~\cite{transformer} mechanism to fuse the sequential representations of the target node $t$ into a single vector. During the representation fusion, \shortModel~ can identify the effective meta-paths for downstream tasks.
\begin{align}
    Q^{\text{fus}}[t] & = \text{mean}\left(\mathbf{H}^{(0)}\left[t\right]W^{\text{FQ}}\right), \nonumber \\
    K^{\text{fus}}[t] & = \mathbf{H}^{(L)}\left[t\right]W^{\text{FK}}, \nonumber\\
    V^{\text{fus}}[t] & = \mathbf{H}^{(L)}\left[t\right]W^{\text{FV}}, \nonumber \\
    A^{\text{fus}}[t] & = \text{Softmax}\left(\frac{Q^{\text{fus}}[t] {K^{\text{fus}}[t]}^\top}{\sqrt{d}}\right), \nonumber \\
    \mathbf{H}\left[t\right] & = A^{\text{fus}}[t]V^{\text{fus}}[t],
\end{align}
where $\mathbf{H}\left[t\right] \in \mathbb{R}^d$ is the final representation of the target node $t$; $W^{\text{FQ}}$, $W^{\text{FK}}$ and $W^{\text{FV}}$ are all learnable matrices of dimension $d \times d$; $Q^{\text{fus}}[t]$ is generated by original features of target node $t$; $A^{\text{fus}}[t] \in \mathbb{R}^{F^{(l)}_{\tau(i)}}$ stands for the importance of each representation for node $t$, which is also the importance of meta-paths.

Referring to \cite{gat,hgt,rhgnn}, we adopt the multi-head attention mechanism during the message passing and representation fusion. The output of the multi-head attention is concatenated into a $d$-dimensional representation to enhance the stability of the model. In addition, we randomly drop out some fragments of the sequential representation of each node in training loops, which can help the \shortModel~ model learn more meaningful node representations.

\section{Experiments}
\label{sec:exp}
In this section, we evaluate the performance of \shortModel~ by conducting experiments on multiple datasets.

\subsection{Datasets}
We conduct extensive experiments on four widely used datasets from Heterogeneous Graph Benchmark (HGB)~\cite{HGB}\footnote{\url{https://github.com/THUDM/HGB}} and Open Graph Benchmark (OGB)~\cite{ogb}\footnote{\url{https://ogb.stanford.edu/}}. Specifically, three medium-scale datasets, DBLP, IMDB and ACM, are from HGB. A large-scale dataset MAG comes from OGB. Their statistics are shown in Table~\ref{tab:dataset_statistics}.
\begin{itemize}
    \item \textbf{DBLP} is a bibliography website of computer science\footnote{\url{https://www.dblp.org/}}. This dataset contains four types of nodes: \textit{Author}, \textit{Paper}, \textit{Term} and \textit{Venue}. In this data set, models need to predict the research fields of authors.
    \item \textbf{IMDB} is extracted from the Internet Movie Database (IMDb)\footnote{\url{https://www.imdb.com/}}. It contains four types of nodes: \textit{Movie}, \textit{Director}, \textit{Keyword} and \textit{Actor}. Models need to divide the movie into 5 categories: ``Romance'', ``Thriller'', ``Comedy'', ``Action, Drama''.
    \item \textbf{ACM} is also a citation network. It contains four types of nodes: \textit{Paper}, \textit{Author}, \textit{Subject (Conference) } and \textit{Term}. The \textit{Paper} nodes are divided into 3 categories: ''database``, ''wireless communication`` and ``data mining''. The model needs to predict the category the paper belongs to.
    \item \textbf{MAG} is a heterogeneous academic network extracted from the Microsoft Academic Graph\footnote{\url{https://www.microsoft.com/en-us/research/project/microsoft-academic-graph/}}, consisting of \textit{Paper}, \textit{Author}, \textit{Field} and \textit{Institution}. Papers are published in 349 different venues. Each paper is associated with a Word2Vec feature. The model needs to predict the category the paper belongs to. The model needs to predict the venues in which the papers are published.
    
\end{itemize}

\begin{table}
  \caption{Statistics of datasets.}
  \label{tab:dataset_statistics}
  \small
  \setlength{\tabcolsep}{0.9mm}
  \begin{tabular}{crcrccc}
  \toprule
   \makecell[c]{name} & \#Nodes & \makecell[c]{\#Node\\Types} & \#Edges & \makecell[c]{\#Edge\\Types}& Target& \#Classes\\ 
   \midrule
  DBLP          & 26,128               & 4                         & 239,566              & 6                         & author                                                                             & 4                            \\
  IMDB          & 21,420               & 4                         & 86,642               & 6                         & movie                                                                              & 5                            \\
  ACM           & 10,942               & 4                         & 547,872              & 8                         & paper                                                                              & 3                            \\
  MAG      & 1,939,743              & 4                         & 21,111,007            & 4                        & paper                                                                               & 349                            \\
  \bottomrule
  \end{tabular}
\end{table}

\begin{table*}[!htbp]
  \begin{tabular}{cccccccc}
  \hline
                       &           & \multicolumn{2}{c}{DBLP}                  & \multicolumn{2}{c}{IMDB}                  & \multicolumn{2}{c}{ACM}                  \\ \hline
                       &           & macro-f1            & micro-f1            & macro-f1            & micro-f1            & macro-f1            & micro-f1           \\ \hline
  \multirow{4}{*}{\makecell{Metapath-based \\ methods}} & RGCN      & 91.52±0.50          & 92.07±0.50          & 58.85±0.26          & 62.05±0.15          & 91.55±0.74          & 91.41±0.75         \\
                       & HetGNN    & 91.76±0.43          & 92.33±0.41          & 48.25±0.67          & 51.16±0.65          & 85.91±0.25          & 86.05±0.25    \\
                       & HAN       & 91.67±0.49          & 92.05±0.62          & 57.74±0.96          & 64.63±0.58          & 90.89±0.43          & 90.79±0.43    \\
                       & MAGNN     & 93.28±0.51          & 93.76±0.45          & 56.49±3.20          & 64.67±1.67          & 90.88±0.64          & 90.77±0.65    \\ \hline
  \multirow{5}{*}{\makecell{Metapath-free \\ methods}} & RSHN      & 93.34±0.58          & 93.81±0.55          & 59.85±3.21          & 64.22±1.03          & 90.50±1.51          & 90.32±1.54    \\
                       & HetSANN   & 78.55±2.42          & 80.56±1.50          & 49.47±1.21          & 57.68±0.44          & 90.02±0.35          & 89.91±0.37     \\
                       & HGT       & 93.01±0.23          & 93.49±0.25          & 63.00±1.19          & 67.20±0.57          & 91.12±0.76          & 91.00±0.76     \\
                       & HGB       & 94.01±0.24          & 94.46±0.22          & 63.53±1.36          & 67.36±0.57          & 93.42±0.44          & 93.35±0.45     \\
                      & SeHGNN    & \underline{95.06±0.17} & \underline{95.42±0.17}          & \textbf{67.11±0.25}          & \underline{69.17±0.43}          & \underline{94.05±0.35}          & \underline{93.98±0.36}     \\ \hline
  \multirow{4}{*}{Ours} & \shortModel & \textbf{96.27±0.24} & \textbf{95.96±0.31} & \underline{66.77±0.24} & \textbf{69.31±0.27}          & \textbf{94.41±0.26}          & \textbf{94.33±0.31}      \\
                       & -w/o seq & 93.79±0.34          & 93.51±0.38          & 64.32±0.56          & 67.04±0.62          & 92.44±0.67          & 92.17±0.72     \\
                       & -w/o fus & 95.59±0.14           & 95.92±0.13          & 65.01±0.37 & 67.43±0.32          & 93.21±0.48          & 93.20±0.50 \\
                       & -w/o rel & 95.49±0.23           & 95.64±0.18          & 64.78±0.41 & 69.09±0.39          & 93.76
                       ±0.43          & 93.67±0.46 \\ \hline
  \end{tabular}
  \caption{Experiment results on the three datasets from the HGB benchmark. The best results are in bold, and the second-best results are underlined.}
  \label{tab:result_on_midlle_dataset}
  \end{table*}

\begin{table}[!t]
  \begin{tabular}{lcc}
  \hline
  Methods               & Validation accuracy & Test accuracy       \\ \hline
  RGCN                  & 48.35±0.36          & 47.37±0.48          \\
  HGT                   & 49.89±0.47          & 49.27±0.61          \\
  NARS                  & 51.85±0.08          & 50.88±0.12          \\
  SAGN                  & 52.25±0.30          & 51.17±0.32          \\
  GAMLP                 & 53.23±0.23          & 51.63±0.22          \\ \hline
  HGT+emb               & 51.24±0.46          & 49.82±0.13          \\
  NARS+emb              & 53.72±0.09          & 52.40±0.16          \\
  GAMLP+emb             & 55.48±0.08          & 53.96±0.18          \\
  SAGN+emb+ms           & 55.91±0.17          & 54.40±0.15          \\
  GAMLP+emb+ms          & 57.02±0.41          & 55.90±0.27          \\
  SeHGNN+emb            & 56.56±0.07          & 54.78±0.17          \\
  SeHGNN+emb+ms         & \underline{59.17±0.09} & \underline{57.19±0.12}          \\ \hline
  \shortModel+emb       & 56.93±0.11         & 55.27±0.34          \\
  \shortModel+emb+ms    & \textbf{59.21±0.08} & \textbf{57.76±0.26} \\\hline
  \end{tabular}
  \caption{Experiment results on the large-scale dataset MAG, where ``emb'' means using extra embeddings and ``ms'' means using multi-stage training. The best results are in bold, and the second-best results are underlined.}
  \label{tab:cls_result}
\end{table}

\subsection{Results Analysis}
\subsubsection{Results on HGB Benchmark}
Table~\ref{tab:cls_result} shows the results of \shortModel~ on the three datasets compared to the baselines in the HGB benchmark.
Baselines are divided into two categories: meta-path-based methods and meta-path-free methods.
Meta-path based methods include RGCN~\cite{rgcn}, HetGNN~\cite{HetGNN}, HAN~\cite{han} and MAGNN~\cite{magnn}.
The meta-path-free methods are RSHN~\cite{RSHN}, HetSANN~\cite{HetSANN}, HGT~\cite{hgt}, HGB~\cite{HGB} and SeHGNN~\cite{SeHGNN}. The results of the baselines are from HGB and their original papers.
As shown in Table~\ref{tab:cls_result}, our proposed method achieves the best performance on ACM and DBLP datasets. In detail, \shortModel~ gains improvement beyond the best baseline on macro-f1 by (1.2\%, 0.4\%) and on mirco-f1 by (0.5\%, 0.4\%), respectively. On the IMDB dataset, our method achieves the best micro f1 scores and the second-best macro f1 scores.
The performance difference between  IMDB and the other two datasets may be due to the following two reasons: (1) Domain difference: DBLP and ACM are datasets in the academic domain while IMDB comes from the film domain. (2) Task difference: IMDB is a multiple-label classification task, but ACM and DBLP are not.

\subsubsection{Results on OGB-MAG} 
Since some types of nodes in the MAG dataset have no initial features, existing methods usually utilize unsupervised representation methods to generate node embeddings (abbreviated as emb) as initial features. For a fair comparison, we also use the unsupervised representation learning method (ComplEx~\cite{complex}) to generate node embeddings. In addition, some baseline methods on the list also adopt multi-stage learning~\cite{ms1,ms2,ms3} (abbreviated as ms) tricks to improve the generalization ability of the model. Therefore, we also explored the performance of \shortModel~ under the multi-stage training.

As shown in Table~\ref{tab:cls_result}, \shortModel~ achieves the best performance compared to the baseline on the ogb leaderboard~\footnote{\url{https://ogb.stanford.edu/docs/leader_nodeprop/\#ogbn-mag}}. 
It shows that our method can not only mine information in heterogeneous graphs 
 more effectively, but also reflect good scalability to be applied to large-scale graphs.

\subsection{Ablation Study}
One of the core contributing components in \shortModel~is to explore how to effectively exploit the structural information in heterogeneous graphs. So we design three variants of our model to verify their effects, namely \textbf{\shortModel~ w/o seq}, \textbf{\shortModel~ w/o fus}, and \textbf{\shortModel~ w/o rel}.
The performance of these variants on the HGB dataset is shown in Table~\ref{tab:result_on_midlle_dataset}.
The details of these variants are as follows:
\begin{itemize}
    \item \textbf{\shortModel~ w/o seq}. It does not use the sequential node representation. After each layer of graph convolution, multiple node representations from different relationships are aggregated into a vector representation by the mean operation. Finally, the \textbf{\shortModel~ w/o seq} concatenates the output of each graph convolutional layer as the final output for the downstream tasks. 
    Comparing \textbf{\shortModel~ w/o seq} and \shortModel, it can be found that after introducing sequential node representation, the performance of the model can be significantly improved. It proves that sequential node representations indeed retain richer and more effective node information.
    \item \textbf{\shortModel~ w/o fus}. It works on the final representation of the node, in which it drops the heterogeneous representation fusion module, instead using the average representation sequence output sent by the last layer of \shortModel. Comparing \textbf{\shortModel~ w/o fus} and \shortModel, it can be found that the performance decreases after removing the heterogeneous fusion module. It illustrates the importance of recognizing the most contributing meta-path.
    \item \textbf{\shortModel~ w/o rel}. It does not add relationship-aware encoding when updating the node representation, which is introduced in equation~\ref{eq:rel_enc}, section~\ref{node_update}. As shown in Table~\ref{tab:cls_result}, \shortModel~ performs better than \textbf{\shortModel~ w/o rel} on all datasets. It verifies the relation-distinguishing ability of \shortModel. 
\end{itemize}

\subsection{Experiment Setup Detail}
\label{sec:cls_setup}
We use the PyTorch Geometric framework 2.0 \footnote{\url{https://www.pyg.org/}} to implement the \shortModel. The source code is available at \url{https://github.com/nobrowning/SEQ_HGNN}. 
We set the node embedding dimension $d= 512$, and the number of attention heads to 8.
The number of layers $L$ is set to 2 on the DBLP, IMDB and MAG datasets and to 3 on the ACM dataset.
During the training process, we set the dropout rate to 0.5, and the maximum epoch to 150. We use the AdamW optimizer~\cite{adamw} with a maximum learning rate of 0.0005 and tune the learning rate using the OneCycleLR strategy~\cite{OneCycleLR}.
For DBLP, ACM, and IMDB datasets, we use full batch training. For the large-scale dataset MAG, we use the HGTLoader\footnote{\url{https://pytorch-geometric.readthedocs.io/en/latest/modules/loader.html}} subgraph sampling strategy~\cite{hgt}, setting the batch size to 256, sampling depth to 3, sample number to 1800. We iterate 250 batches in each epoch.

The results of the baselines in Table~\ref{tab:result_on_midlle_dataset} and  Table~\ref{tab:cls_result} mainly come from previous works~\cite{HGB, SeHGNN}. All experiments can be conducted on a Linux machine with Intel(R) Core(R) i7 8700 CPU, 32G RAM, and a single NVIDIA GeForce RTX 3090 GPU.

\subsection{Training Efficiency}
In \shortModel, sequential node representations are computed in parallel. Therefore, \shortModel~ achieves decent computational efficiency.
To further investigate the computational efficiency of \shortModel, we conduct experiments to compare the training time of \shortModel~ with a state-of-the-art baseline, i.e., SeHGNN.

\begin{figure}[t]
  \centering
  \includegraphics[width=0.8\columnwidth]{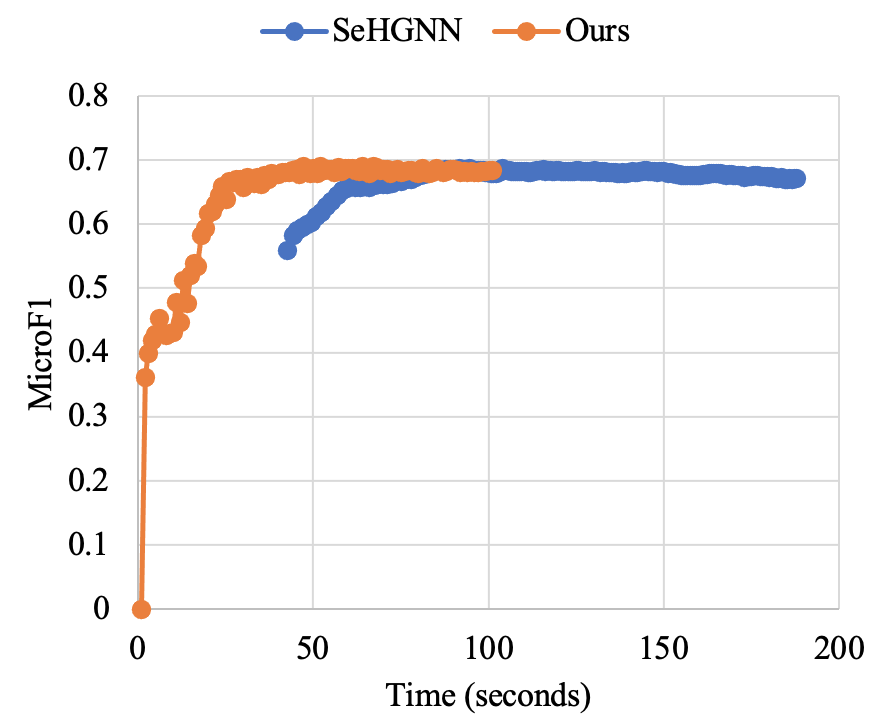}
  \caption{The comparison of training efficiency.}
  \label{fig:time}
  \Description{The comparison of training efficiency.}
\end{figure}

To achieve a fair comparison, we subject all models to the same accuracy performance validation --- making a test on the test set every one train epoch. The variation of test accuracy of the models with training time is shown in Figure~\ref{fig:time}.

As shown in Figure~\ref{fig:time}, \shortModel~ performs the highest accuracy within the least training time. It verifies that \shortModel~ has good computational efficiency when dealing with heterogeneous graphs. 
As a comparison, 
the baseline (SeHGNN) outputs nothing within 42 seconds of starting training. The reason is that SeHGNN cannot directly learn node representations on heterogeneous graphs. It requires a message-passing step before node representation generation. In the message passing step, SeHGNN collects the features of neighbor nodes of the target on all meta-paths. Therefore, the messaging step shows a high time-consuming.

\begin{figure}
  \centering
  \includegraphics[width=1\columnwidth]{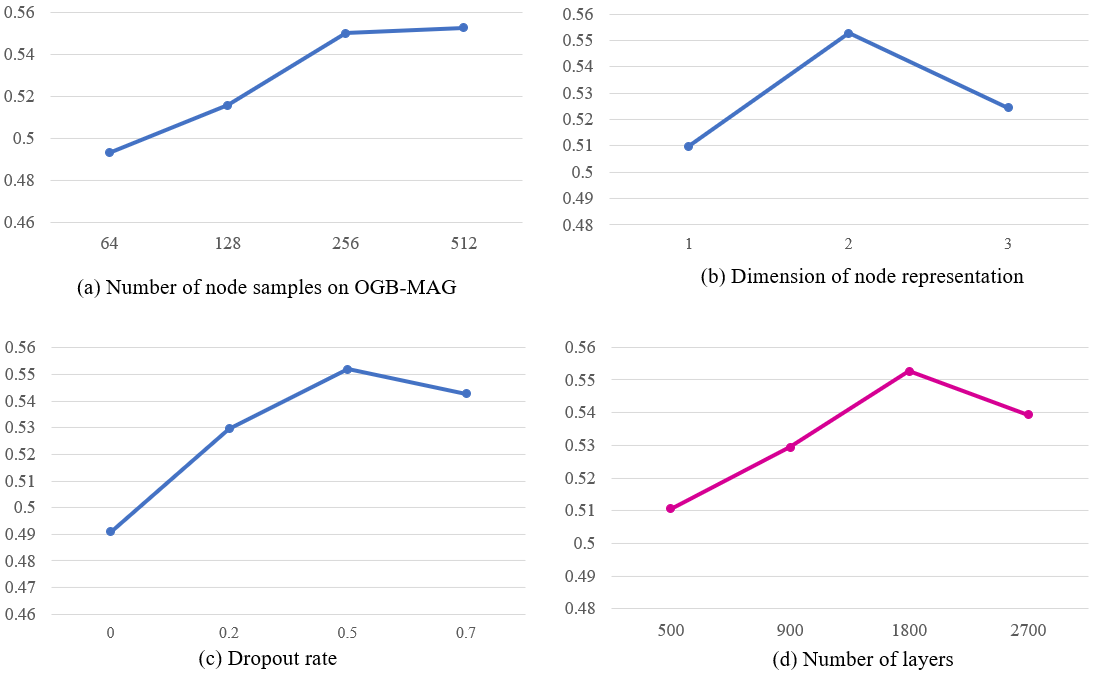}
  \caption{Parameter Sensitivity of \shortModel.}
  \label{fig:params}
  \Description{Parameter Sensitivity of \shortModel.}
\end{figure}

\begin{figure*} 
  \centering
  \subfloat[DBLP.\label{fig:case_dblp}]{%
       \includegraphics[width=0.5\columnwidth]{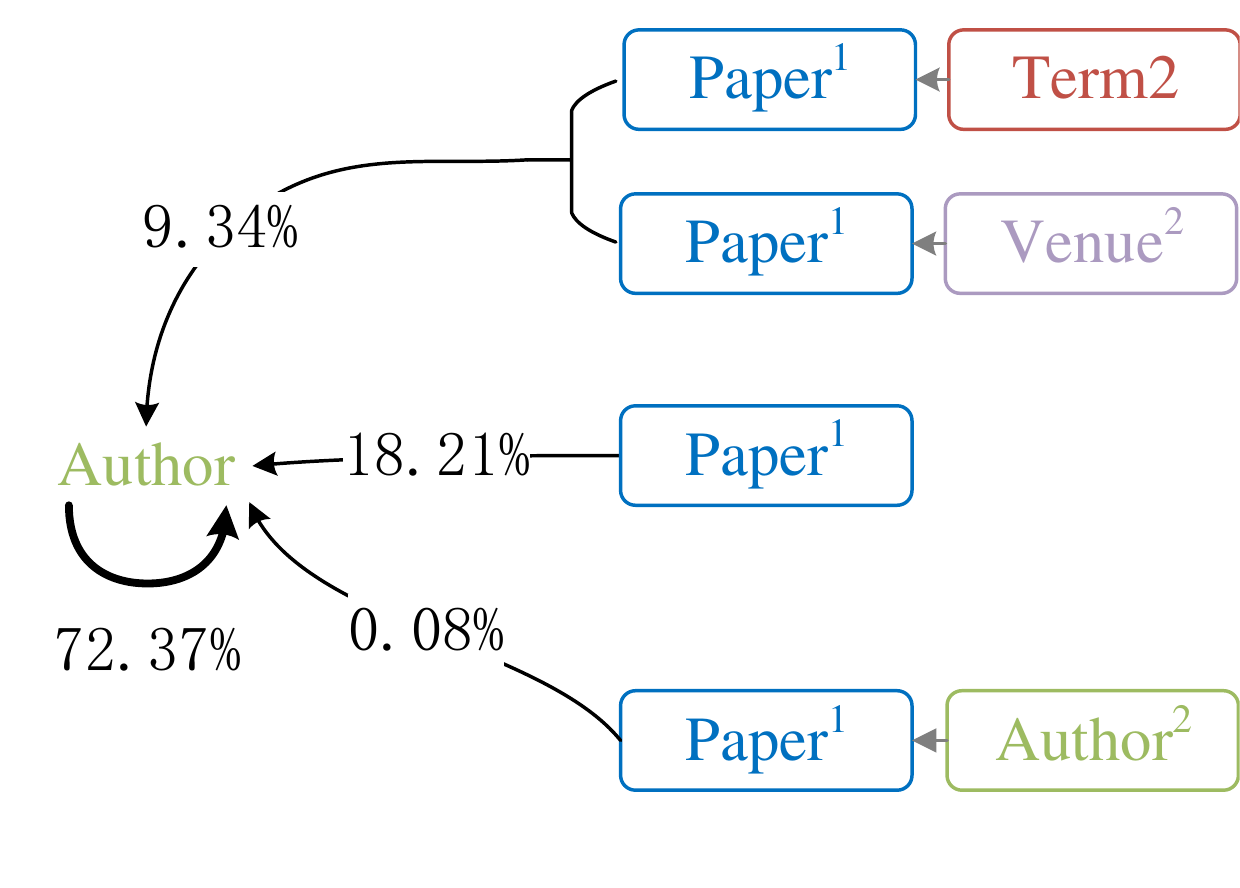}}
    \hfill
  \subfloat[IMDB.\label{fig:case_imdb}]{%
        \includegraphics[width=0.5\columnwidth]{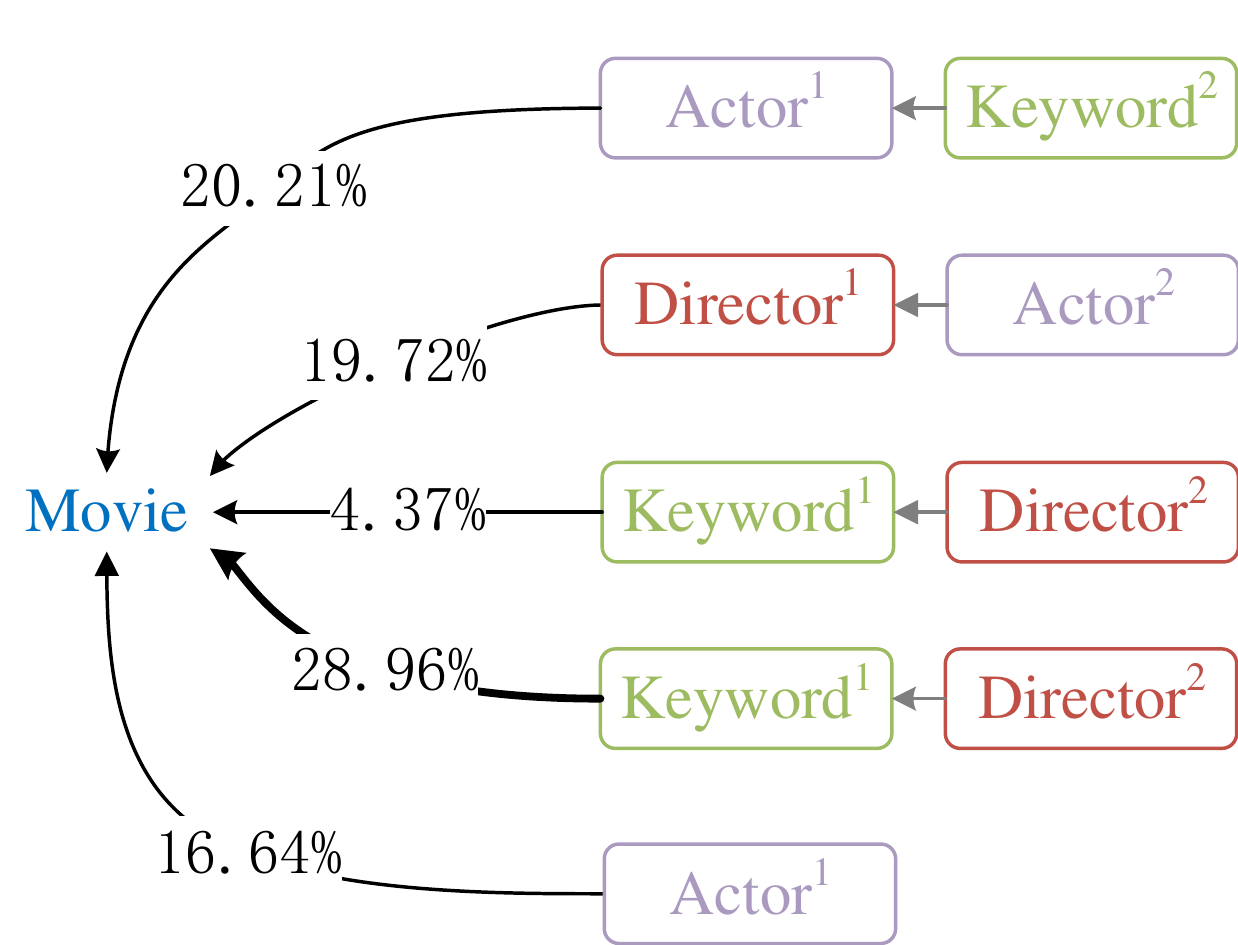}}
  \subfloat[ACM.\label{fig:case_acm}]{%
        \includegraphics[width=0.5\columnwidth]{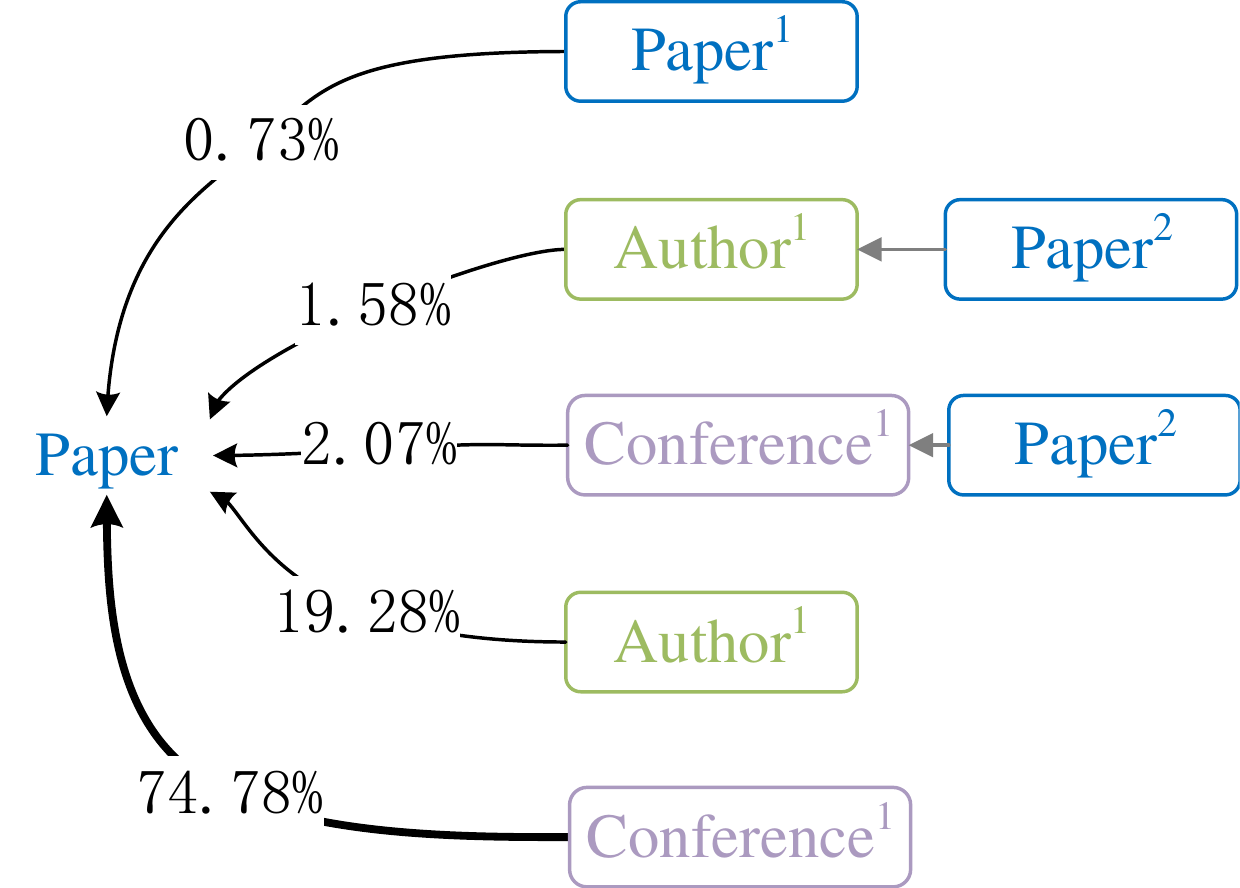}}
  \subfloat[MAG.\label{fig:case_mag}]{%
        \includegraphics[width=0.5\columnwidth]{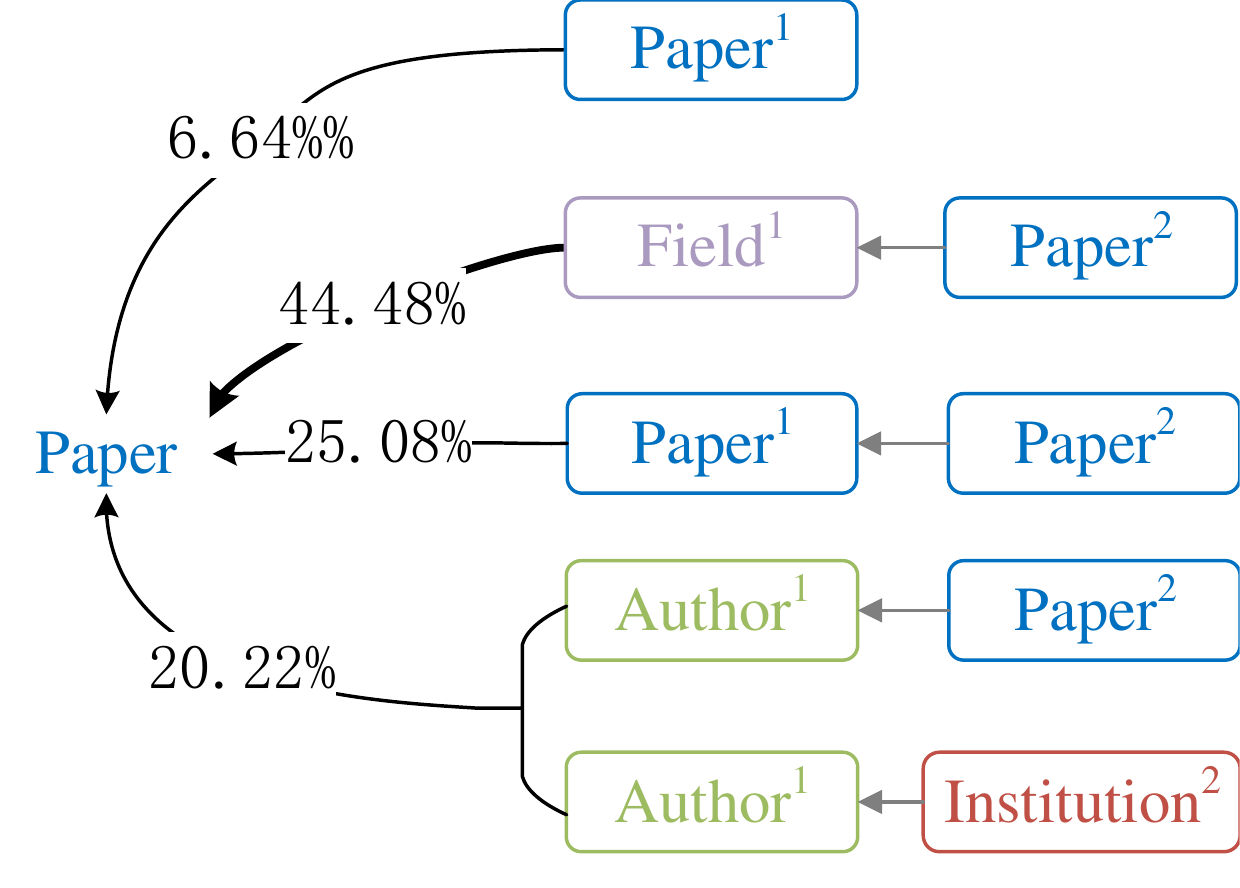}}
  \caption{Visualization of the significant meta-paths for representing the target nodes (Author, Movie, Paper, Paper) in respective datasets (DBLP, IMDB, ACM, MAG). 
  In the figures, the nodes with superscripts $1$ and $2$ represent the direct neighbors and the second-order neighbors of the target node, respectively.
  }
  \label{fig:person_representation_case}
  \Description{Visualization of the significant meta-paths for representing the target nodes (Author, Movie, Paper, Paper) in respective datasets (DBLP, IMDB, ACM, MAG). 
  In the figures, the nodes with superscripts $1$ and $2$ represent the direct neighbors and the second-order neighbors of the target node, respectively.}
\end{figure*}

  \subsection{Parameter Sensitivity Analysis}
  We study the sensitivity analysis of parameters in \shortModel.
  Specifically, we conduct experiments on the large-scale dataset OGB-MAG to explore the influence of the number of layers, the dropout rate, and the dimension of node representation.
  Since the model needs to conduct a sub-graph sampling on the large-scale dataset, we also explore the influence of sampling node numbers.
  To simplify the evaluation process, we opted not to employ a multi-stage training strategy in the parameter sensitivity experiment.
  The results are shown in Figure~\ref{fig:params}, where each subfigure shows the accuracy of classification on the y-axis and hyperparameters on the x-axis.
  
  \subsubsection{Number of node samples.}
  Since \shortModel~ uses HGT-Loader for sampling sub-graphs in the node classification task, we explore the effect of node sampling number on the performance of \shortModel. As shown in Figure~\ref{fig:params} (a), \shortModel~ achieves the best performance when the number of samples is set as 1800.
  
  \subsubsection{Dimension of node representation.}
  We report the experimental result varied with the dimension of node representation in Figure~\ref{fig:params} (b). It can be seen that as the dimension increases, the performance of \shortModel~ gradually increases. After the dimension is higher than 256, the performance improvement slows down.
  
  \subsubsection{Dropout rate.}
  We adjust the dropout rate during the model training and report the results in Figure~\ref{fig:params} (c).
  We can observe that \shortModel~ performs best when the dropout rate is 0.5. A high dropout rate would lead to underfitting and poor performance, while a low dropout rate may lead to overfitting.
  
  \subsubsection{Number of layers.}
  We explore the performance of our model while stacking from 1 to 3 \shortModel~ Layers. The experimental results are shown in Figure~\ref{fig:params} (d). It can be seen that \shortModel~ achieves the best performance when it is stacked with 2 layers. 
  On this basis, the performance of \shortModel~ becomes worse when more layers are stacked. This may be caused by over-smoothing issues.

\subsection{Visualization of Effective Meta-Paths}
As mentioned in Section~\ref{sec:feat_fus}, $A^{\text{fus}}$ in the Heterogeneous Representation Fusion module indicates the importance of different representations of a node, i.e., the importance of a node on different meta-paths.
To visualize how the heterogeneous fusion module of \shortModel~ identifies the most contributing meta-paths, we present the effective meta-paths in node representation learning on DBLP, IMDB, ACM and MAG datasets, respectively.
The most important meta-paths for these target node representations are shown in Figure~\ref{fig:person_representation_case}. It is noteworthy that our model can individually identify the significant metapaths characterizing each node. In order to simplify the visualization, we aggregate the propagation path weights of nodes by node type in Figure~\ref{fig:person_representation_case}. Due to the large number of meta-paths, here, we only show the top five important paths in each sub-figure.

Comparing the four sub-figure in Figure~\ref{fig:person_representation_case}, we can find that the important paths for distinct nodes are obviously different. It verifies that the \shortModel~ can estimate the path importance separately for different nodes, rather than treat them equally.

In sub-figure (a), we can observe that the self-loop of the target node (\textit{Author}) has a high weight (72.37\%). It reveals that in the ACM dataset, the representation of the \textit{Author} node mainly depends on its own attributes rather than the structural information in the graph. 
In contrast, the information of the target node (\textit{Movie}) in sub-figure (b) mainly comes from its neighbor nodes.
The target node types in sub-figure (c) and sub-figure (d) are both \textit{Paper}. However, there is a significant difference between sub-figure (c) and sub-figure (d): the most important meta-path in sub-figure (c) is ``\textit{Paper-Conference}'', while the information of the target node in sub-figure (d) mostly comes from the meta-paths related to \textit{Paper}, such as ``\textit{Paper-Field-Paper}'', ``\textit{Paper-Paper-Paper}'', ``\textit{Paper-Author-Paper}'', etc. The difference between sub-figure (c) and sub-figure (d) may be mainly caused by their downstream tasks. Specifically, the task of sub-figure (c)  is to predict the field of the paper while the task of sub-figure (d)  is to predict the journal where the paper is published. This indicates that our model can utilize different aspects of graphs according to different downstream task demands. By mining important propagation paths, the model can provide deep insights and interpretability into the real-world application scenarios.

\section{Conclusion}
In this paper, we proposed a novel heterogeneous graph neural network with sequential node representation, namely \shortModel.
To avoid the information loss caused by the single vector node representation, we first design a sequential node representation learning mechanism to represent each node as a sequence of meta-path representations during the node message passing.
Then we propose a heterogeneous representation fusion module, empowering \shortModel~ to identify important meta-paths and aggregate their representations into a compact one.
Third, we conducted extensive experiments on four widely-used datasets from open benchmarks and clearly validated the effectiveness of our model.
Finally, we visualized and analyzed effective meta-path paths in different datasets, and verified that \shortModel~ can provide deep insights into the heterogeneous graphs.

\balance

\begin{acks}
This research work is supported by the National Key Research and Development Program of China under Grant No. 2019YFA0707204, the National Natural Science Foundation of China under Grant Nos. 62176014, 62276015, the Fundamental Research Funds for the Central Universities.
\end{acks}

\bibliographystyle{ACM-Reference-Format}
\bibliography{sample-base}

\end{document}